\documentclass[5pt]{article}

\usepackage[letterpaper]{geometry}
\usepackage{spconf,amsmath,epsfig}
\usepackage{enumitem}
\usepackage{times}
\usepackage{helvet}
\usepackage{courier}
\usepackage{caption}
\usepackage{epsfig}
\usepackage{graphicx}
\usepackage{amsmath}
\usepackage{amssymb}
\usepackage{multirow}
\usepackage{epstopdf}
\usepackage{graphicx}

\usepackage{fancyhdr}
\begin{document}\sloppy

\def\x{{\mathbf x}}
\def\L{{\cal L}}

\title{Cross-modal Deep Metric Learning with Multi-task Regularization}
%
\name{Xin Huang and Yuxin Peng*\thanks{*Corresponding author.}}
\address{Institute of Computer Science and Technology, Peking University \\
Beijing 100871, China \\
\{huangxin\_14, pengyuxin\}@pku.edu.cn}

%
%
%

\maketitle

\begin{abstract}
DNN-based cross-modal retrieval has become a research hotspot, by which users can search results across various modalities like image and text. However, existing methods mainly focus on the pairwise correlation and reconstruction error of labeled data. They ignore the semantically similar and dissimilar constraints between different modalities, and cannot take advantage of unlabeled data. This paper proposes Cross-modal Deep Metric Learning with Multi-task Regularization (CDMLMR), which integrates quadruplet ranking loss and semi-supervised contrastive loss for modeling cross-modal semantic similarity in a unified multi-task learning architecture. The quadruplet ranking loss can model the semantically similar and dissimilar constraints to preserve cross-modal relative similarity ranking information. The semi-supervised contrastive loss is able to maximize the semantic similarity on both labeled and unlabeled data. Compared to the existing methods, CDMLMR exploits not only the similarity ranking information but also unlabeled cross-modal data, and thus boosts cross-modal retrieval accuracy.
\end{abstract}
\section{Introduction}
Nowadays, multimedia retrieval is increasingly important for data management and utilization, and has been a research hotspot for a long time. However, most of the existing methods are for single-modal retrieval, and can only measure the similarity between data of the same single modality. 
Different modalities are different views of semantics. An image of flying bird and a text description of bird have the same semantic of ``bird". So they describe the same semantics through two different views, and are similar to each other in the semantic level. Modeling the similarities among different modalities is important for better understanding the multimedia data, and also for multimedia applications on the Internet.

Cross-modal similarity learning focuses on exploiting semantic correlation among multiple modalities like image and text. The existing methods mainly project data of different modalities into one common space and then get shared representations for them. So the cross-modal similarity can be directly measured by distance computing. For example, canonical correlation analysis (CCA)
 \cite{HotelingBiometrika36RelationBetweenTwoVariates} can learn a common space maximizing correlation between data with two modalities. There are many CCA-based methods as \cite{RasiwasiaMM10SemanticCCA,DBLP:journals/ijcv/GongKIL14}. Cross-modal Factor Analysis (CFA) approach \cite{LiMM03CFA} aims to minimize Frobenius norm between pairwise data in the common space. Besides, Zhai et al. propose Joint Representation Learning (JRL) \cite{ZhaiTCSVT2014JRL}, which can model pairwise correlation and semantic information jointly in a unified graph-based framework. Kang et al. propose Local Group based Consistent Feature Learning (LGCFL) \cite{DBLP:journals/tmm/KangXLXP15}, which adopts a local group-based priori to learn basis matrices of different modalities.
 Hua et al. \cite{DBLP:journals/tmm/HuaWLCH16} propose to first build semantic hierarchy with content and ontology similarities, and then learn a set of local linear projections and probabilistic membership functions for local expert aggregation.
 However, the above methods mostly perform shared representation projection by linear functions, which is insufficient for the complex cross-modal correlation.

Inspired by the improvement of deep neural network (DNN) in single-modal retrieval, researchers have also attempted to apply DNN to cross-modal similarity measure. For instance, the input of different modalities can be converted to shared representations through a shared code layer as \cite{ngiam32011multimodal,srivastava42012multimodal,DBLP:conf/ijcai/PengHQ16}. Also, 
there are some cross-modal deep architectures consisting of two linked deep encodings such as Deep CCA \cite{DBLP:conf/icml/AndrewABL13,DBLP:conf/cvpr/YanM15} and Corr-AE \cite{feng12014cross}.
The above methods mainly focus on the pairwise correlation and reconstruction error of labeled multimodal data. However, they ignore the semantically similar and dissimilar constraints between different modalities, which can provide similarity ranking information for better semantically discriminative ability. Unlabeled data should also be taken into consideration, which can increase the diversity of training data and boost the accuracy of shared representation learning.

\begin{figure*}[t]
  \centering
\begin{minipage}[c]{0.95\linewidth}
\centering
  \includegraphics[width=\textwidth]{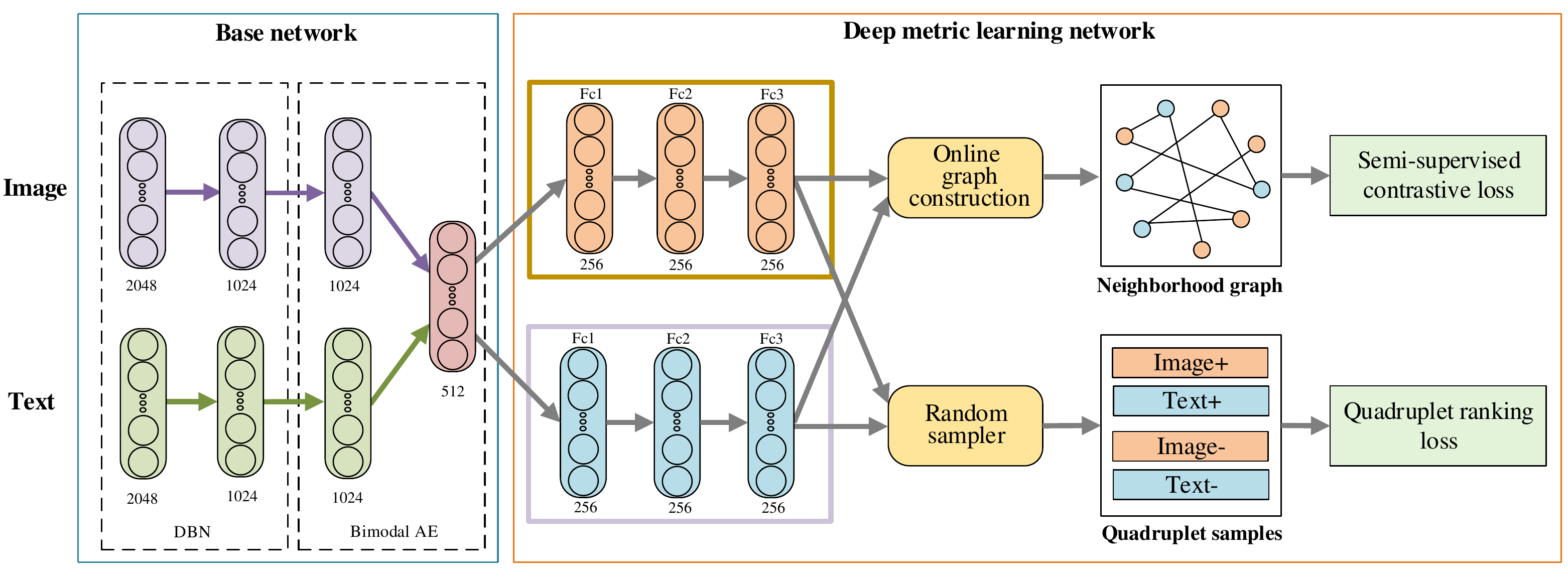}
\end{minipage}%
\caption{An overview of our CDMLMR model.}\label{fig:network}
\vspace{-3mm}
\end{figure*}

For addressing the above problems, this paper proposes Cross-modal Deep Metric Learning with Multi-task Regularization (CDMLMR), which integrates quadruplet ranking loss and semi-supervised contrastive loss for modeling cross-modal semantic similarity in a unified  metric learning architecture. 
On the one hand, triplet network has been proposed for single-modal metric learning \cite{DBLP:conf/cvpr/WangSLRWPCW14,DBLP:conf/cvpr/LaiPLY15}, which can model the relative similarity of images. CDMLMR extends the single-modal triplet network to cross-modal quadruplet network, 
which can preserve relative similarity ranking information between different modalities. On the other hand, the semi-supervised contrastive loss can preserve the similarity information on both labeled and unlabeled data by maximizing the semantic correlation with an online graph construction strategy. 
The two loss functions can be integrated into one unified multi-task network and optimized simultaneously inspired by \cite{DBLP:conf/nips/RenHGS15}.  By doing so, CDMLMR can capture fully the useful yet intrinsic hints for cross-modal similarity measure and improve the retrieval accuracy.
Experiment results show that CDMLMR achieves better performance comparing with 8 state-of-the-art methods on 2 datasets: Wikipedia \cite{RasiwasiaMM10SemanticCCA} and NUS-WIDE-10k \cite{NUSWIDE}.

\section{Cross-modal Deep Metric Learning with Multi-task Regularization}

To perform cross-modal deep metric learning, a base network with Deep Belief Network (DBN) and Bimodal Autoencoders (Bimodal AE) \cite{ngiam32011multimodal} will first be used for the feature of different modalities, from which we can get the shallow cross-modal shared representations. 
Then as shown in Figure \ref{fig:network}, CDMLMR model integrates two loss functions for modeling the cross-modal semantic similarity in a unified optimization deep network, which are semi-supervised contrastive loss and quadruplet ranking loss.
The network is trained by simultaneous optimization based on these loss functions to get the final semantically discriminative shared representations.

Formally, the multimodal dataset contains both the labeled data and unlabeled data. $D_{L}^{(i)}=\left \{x_{p}^{(i)},y_{p}^{(i)}\right \}_{p=1}^{M}$ denotes the labeled image data, here the $p$-th image data is denoted as $x_{p}^{(i)}\in \mathbb{R}^{d^{(i)}}$, $M$ is the number of the labeled image data, the dimensional number of the image feature is $d^{(i)}$, and $y_{p}^{(i)}$ is the corresponding label of $x_{p}^{(i)}$. The unlabeled image data is denoted as $D_{U}^{(i)}=\left \{x_{p}^{(i)}\right \}_{p=M+1}^{N}$, where $N$ is the total number of the image data. And the text data is represented as $D_{L}^{(t)}=\left \{x_{p}^{(t)},y_{p}^{(t)}\right \}_{p=1}^{M}$ and $D_{U}^{(t)}=\left \{x_{p}^{(t)}\right \}_{p=M+1}^{N}$ which is similar to the image data. It should be noted that $p$-th image and text, i.e., $x_{p}^{(i)}$ and $x_{p}^{(t)}$, have pairwise correspondence.

\subsection{The Base Network}

The base network can convert data from different modalities to representations of the same dimensional number, and these shallow shared representations will be used as input of the deep metric learning network.
We first employ a separate two-layer DBN to model each modality. To model the distribution over the image feature $\left \{x_{p}^{(i)}\right \}$, the Gaussian Restricted Boltzmann Machine (RBM) is used, which is an undirected graphical model having visible units $v$ connected to the hidden units $h$. And for text feature $\left \{x_{p}^{(t)}\right \}$, Replicated Softmax is used to model the distribution over them. The probability that each DBN model assigns to the input feature is defined as follows:
\begin{small}
\begin{gather}
P(v_{i})=\sum_{h_i^{(1)},h_i^{(2)}}P(h_i^{(2)},h_i^{(1)})P(v_{i}|h_i^{(1)}) \\
P(v_{t})=\sum_{h_t^{(1)},h_t^{(2)}}P(h_t^{(2)},h_t^{(1)})P(v_{t}|h_t^{(1)})
\end{gather}
\end{small}

The outputs of the two DBN are denoted as $Y^{(i)}$ and $Y^{(t)}$. %
Then we use Bimodal AE to get the shallow shared representations with the same dimension. Bimodal AE has the ability to reconstruct both two modalities by minimizing the reconstruction error between the input and the reconstruction representations at the reconstruction layers.
The shallow shared representations obtained from the middle layer of Bimodal AE are denoted as $S^{(i)}$ and $S^{(t)}$.

\subsection{Multi-task Regularization}

As shown in Figure \ref{fig:network}, our CDMLMR model has two pathways of three fully-connected layers for each modality separately, taking the shallow shared representations $S^{(i)}$ and $S^{(t)}$ obtained from the base network as input. At the top of the two pathways network, multiple loss branches are embedded with a fully-connected layer using sigmoid nonlinearity, which integrates the semi-supervised and quadruplet ranking regularization in a unified optimization deep network.
For image, a batch data $X^{(i)}$ for each iteration consist of the labeled images $X_{L}^{(i)}=\left \{s_{p}^{(i)},y_{p}^{(i)}\right \}_{p=1}^{m}$ and the unlabeled images $X_{U}^{(i)}=\left \{s_{p}^{(i)}\right \}_{p=m+1}^{n}$, where $n$ is the total number of the image data in a mini-batch, and $m$ of them are labeled images. Similarly, we have  $X_{L}^{(t)}$ and $X_{U}^{(t)}$. The outputs from the two pathways are separately denoted as $f(s_{p}^{(i)})$ and $g(s_{p}^{(t)})$, where $f(.)$ denotes the image mapping and $g(.)$ denotes the text mapping.

{\textbf {Semi-supervised Contrastive Loss}}: In CDMLMR, the semi-supervised contrastive loss is proposed to preserve the similarity information of both labeled and unlabeled data. The basic idea is similar image/text pairs should have similar shared representations, and vice versa. Here a pair of data are ``similar" if they are close to each other in the shallow shared representation space, or from the same semantic class. For modeling such semi-supervised information, a neighborhood graph $G=(V,E)$ is constructed, where the vertices $V$ represent both image and text data, and the edges $E$ represent the cross-modal similarity matrix between image and text data, which is denoted as $C$ for the labeled image/text pairs and $A$ for the unlabeled image/text pairs. 
For the labeled image $s_{p}^{(i)}$ and labeled text $s_{q}^{(t)}$, the similarity matrix $C$ is constructed based on labels as follows:
\begin{small}
 \begin{gather}
 C(p,q)=\begin{cases}
 1 & : y_{p}^{(i)}=y_{q}^{(t)}\\ 
 0 & : y_{p}^{(i)}\neq y_{q}^{(t)} 
 \end{cases}
 \end{gather}
 \end{small}As for unlabeled image/text pairs which mean that at least one of the image and text data in pair is unlabeled, we analyze the \emph{k-nearest-neighbors} $NN_{k}(s_{p}^{(i)})$ and $NN_{k}(s_{q}^{(t)})$ for each image $s_{p}^{(i)}$ and text $s_{q}^{(t)}$. Instead of constructing the graph offline on all the data which is much time-consuming, an online graph construction strategy is proposed to generate the cross-modal similarity matrix for the unlabeled image/text pairs within a mini-batch. 
 So the similarity matrix $A$ for the unlabeled image/text pairs is defined as follows:
 \begin{small}
 \begin{gather}
 A(p,q)=\begin{cases}
 1 & : s_{p}^{(i)}\in NN_{k}(s_{q}^{(t)})\vee s_{q}^{(t)}\in NN_{k}(s_{p}^{(i)})\\ 
 0 & : s_{p}^{(i)}\notin NN_{k}(s_{q}^{(t)})\wedge s_{q}^{(t)}\notin NN_{k}(s_{p}^{(i)}) 
 \end{cases}
 \end{gather}
  \end{small}For maximizing the semantic correlation, we expect the similar image/text pairs to have smaller distance and the dissimilar image/text pairs to have larger distance. Thus, a contrastive loss for the labeled image/text pairs to capture the similarity information is designed as follows:
 \begin{scriptsize}
 \begin{gather}
 L_{c}(s_{p}^{(i)},s_{q}^{(t)})=\begin{cases}
 \left \| f(s_{p}^{(i)})-g(s_{q}^{(t)}) \right \|^{2} & ,C(p,q)=1 \\ 
 max(0,\alpha-\left \| f(s_{p}^{(i)})-g(s_{q}^{(t)}) \right \|^{2}) & ,C(p,q)=0 
 \end{cases} \label{equ:ls}
 \end{gather}
   \end{scriptsize}and for capturing the adjacency neighbors information, the contrastive loss between unlabeled image/text pairs is defined as follows:
  \begin{scriptsize}
  \begin{gather}
 L_{a}(s_{p}^{(i)},s_{q}^{(t)})=\begin{cases}
 \left \| f(s_{p}^{(i)})-g(s_{q}^{(t)}) \right \|^{2} & ,A(p,q)=1 \\ 
 max(0,\alpha-\left \| f(s_{p}^{(i)})-g(s_{q}^{(t)}) \right \|^{2}) & ,A(p,q)=0 
 \end{cases} \label{equ:la}
 \end{gather}
    \end{scriptsize}where $s_{p}^{(i)}$ and $s_{q}^{(t)}$ are for input image and text data obtained from the base network respectively, and $\alpha$ is the margin parameter. Combining the above two loss functions, finally we get the semi-supervised contrastive loss function as follows:
   \begin{small}
 \begin{gather}
 L=\sum_{j,k=1}^{m}L_{c}(s_{j}^{(i)},s_{k}^{(t)})+ \sum_{j,k=m+1}^{n}L_{a}(s_{j}^{(i)},s_{k}^{(t)})
 \label{equ:l}
 \end{gather}
 \end{small}For balancing the number of similar and dissimilar pairs, we randomly select a similar pair and a dissimilar pair for each image $s_{p}^{(i)}$ or text $s_{q}^{(t)}$ for training. By minimizing the above loss function, we can preserve the similarity information on both the labeled and unlabeled data.
{\textbf {Quadruplet Ranking Loss}}: Inspired by the idea of preserving the relative similarity in the triplet network as \cite{DBLP:conf/cvpr/WangSLRWPCW14}, the cross-modal quadruplet ranking loss is designed for further modeling the cross-modal relative similarity with a sample layer to generate the quadruplet samples from the output of the separate two-pathway network. The quadruplet samples are organized into the form of ($s^{I+},s^{T+},s^{I-},s^{T-}$) and satisfy the following two relative similarity constraints: (1) The text sample $s^{T+}$ is more similar to the image sample $s^{I+}$ than to the image sample $s^{I-}$. (2) The image sample $s^{I+}$ is more similar to the text sample $s^{T+}$ than to the text sample $s^{T-}$. The similarity is according to data labels, so the quadruplet samples are generated only from the labeled data $X_{L}^{(i)}$ and $X_{L}^{(t)}$ in a mini-batch. 
Based on this, the quadruplet ranking loss function is defined as follows:
\begin{small}
\begin{gather}
L(s^{I+},s^{T+},s^{I-},s^{T-}) =max(0, 2\left \| f(s^{I+})-g(s^{T+}) \right \|^{2} \notag \\
-\left \| f(s^{I+})-g(s^{T-}) \right \|^{2} 
-\left \| f(s^{I-})-g(s^{T+}) \right \|^{2}+\beta) \notag \\
\label{equ:loss}
\end{gather}
 \end{small}where $\beta$ is the margin parameter.
By capturing both the between-class and within-class differences between different modalities, the quadruplet network can effectively preserve the cross-modal relative similarity and improve the semantic discriminative ability of shared representations to boost retrieval accuracy.

After network training, we get the mapping function $f(.)$ for the image pathway and $g(.)$ for the text pathway. For both modalities $S^{(i)}$ and $S^{(t)}$, we can calculate $f(S^{(i)})$ and $g(S^{(t)})$ (denoted as $Q^{(i)}$ and $Q^{(t)}$) as the final semantically discriminative shared representations. They can further be used for retrieval by distance measure.

\subsection{Network Training}

CDMLMR involves two loss functions: the quadruplet ranking loss and the semi-supervised contrastive loss. First, we calculate the derivative of the two loss functions separately.
For the semi-supervised contrastive loss, the derivative of the loss function $L_{c}$ in (\ref{equ:ls}) is calculated for each image $p$ and text $q$ as follows:
\begin{small}
\begin{equation}
\begin{split}
& \frac{\partial L_{c}}{\partial f(s_{p}^{(i)})}=\sum_{q=1,C(p,q)=1}^{m}D(s_{p}^{(i)},s_{q}^{(t)}) \\
& -\sum_{q=1,C(p,q)=0}^{m}(\alpha-\left \| D(s_{p}^{(i)},s_{q}^{(t)}) \right \|)\times sgn(D(s_{p}^{(i)},s_{q}^{(t)}))\\
\end{split}
\end{equation}
\begin{equation}
\begin{split}
& \frac{\partial L_{c}}{\partial g(s_{q}^{(t)})}=\sum_{p=1,C(p,q)=1}^{m}D(s_{q}^{(t)},s_{p}^{(i)}) \\
& -\sum_{p=1,C(p,q)=0}^{m}(\alpha-\left \| D(s_{q}^{(t)},s_{p}^{(i)}) \right \|)\times sgn(D(s_{q}^{(t)},s_{p}^{(i)})))\\
\end{split}
\end{equation}
 \end{small}where $D(s_{p}^{(i)},s_{q}^{(t)})=f(s_{p}^{(i)})-g(s_{q}^{(t)})$ and $D(s_{q}^{(t)},s_{p}^{(i)})$ is in opposite.  $\alpha$ is the margin parameter.  Moreover, we can easily calculate the derivative of the $L_{a}$ in (\ref{equ:la}) similar as $L_{c}$ and further calculate the derivative of $L$ in (\ref{equ:l}). Thus, the back-propagation can be applied to update the parameters through the network.

For the quadruplet ranking loss in (\ref{equ:loss}), we calculate the gradients of $f(s^{I+}), g(s^{T+}), f(s^{I-}), g(s^{T-})$ as follows:
\begin{small}
\begin{align}
\frac{\partial L}{\partial i^{+}}=(2i^{+}-4t^{+}+2t^{-})\times C \\
\frac{\partial L}{\partial t^{+}}=(2t^{+}-4i^{+}+2i^{-})\times C  \\
\frac{\partial L}{\partial i^{-}}=(2t^{+}-2i^{-})\times C  \\
\frac{\partial L}{\partial t^{-}}=(2i^{+}-2t^{-})\times C 
\end{align}
 \end{small}where $f(s^{I+}), g(s^{T+}), f(s^{I-}), g(s^{T-})$ is denoted as $i^{+},t^{+},i^{-},t^{-}$. And the parameter $C$ is 1 if
 
\begin{small} 
\begin{equation}
\begin{split}
& 2\left \| f(s^{I+})-g(s^{T+}) \right \|^{2}-\left \| f(s^{I+})-g(s^{T-}) \right \|^{2}  \\
& -\left \| g(s^{T+})-f(s^{I-}) \right \|^{2}+\beta > 0 \\
\end{split}
\end{equation}
 \end{small}otherwise $C=0$. Thus this loss function in (\ref{equ:loss}) could be applied to back propagation in the neural networks. 

After calculating the derivative of the above two loss functions, the gradients of each modality from the fully-connected layers of each loss branch are summed together at the top of the 3 fully-connected layers in the proposed two pathways network for parameter updating.

\begin{table*}[htb!]
\scriptsize
\caption{The MAP scores for \emph{all results}.}
\vspace{-3mm}
\begin{center}
\begin{tabular}{|c|c|c|c|c|c|c|c|c|c|c|c|} 
\hline
Datasets & Task & CCA & CFA & {\begin{tabular}{c} KCCA\\(Poly) \end{tabular}} & {\begin{tabular}{c} KCCA\\(RBF) \end{tabular}} & {\begin{tabular}{c} Bimodal\\AE \end{tabular}} & {\begin{tabular}{c} Multimodal\\DBN \end{tabular}} & Corr-AE & JRL & CMDN & \textbf{CDMLMR}\\
\hline
\multirow{3}{*}{\begin{tabular}{c} Wikipedia \\ Dataset \end{tabular}} &
Image$\rightarrow$Text & 0.124 &0.236 &0.200 & 0.245 & 0.236 & 0.149 & 0.280 & 0.344 & 0.393 & \textbf{0.412} \\
\cline{2-12}
& Text$\rightarrow$Image & 0.120 & 0.211 & 0.185 & 0.219 & 0.208 & 0.150 & 0.242 & 0.277 & 0.325 & \textbf{0.341} \\
\cline{2-12}
& Average & 0.122 & 0.224 & 0.193 & 0.232 & 0.222 & 0.150 & 0.261 & 0.311 & 0.359 & \textbf{0.377} \\
\hline
\multirow{3}{*}{\begin{tabular}{c} NUS-WDIE\\-10k Dataset \end{tabular}} &
Image$\rightarrow$Text & 0.120 &0.211 &0.150 & 0.232 & 0.159 & 0.158 & 0.223 & 0.324 &  0.391 & \textbf{0.405} \\
\cline{2-12}
& Text$\rightarrow$Image & 0.120 & 0.188 & 0.149 & 0.213 & 0.172 & 0.130 & 0.227 & 0.263 & 0.357 & \textbf{0.379} \\
\cline{2-12}
& Average & 0.120 & 0.200 & 0.150 & 0.223 & 0.166 & 0.144 & 0.225 & 0.294 & 0.374 &\textbf{0.392} \\
\hline
\end{tabular} 
\end{center}
\vspace{-3mm}
\label{table:res_wiki_All}
\end{table*}

\begin{table*}[htb!]
\scriptsize
\caption{The MAP scores for \emph{top 50 results}.}
\vspace{-3mm}
\begin{center}
\begin{tabular}{|c|c|c|c|c|c|c|c|c|c|c|c|} 
\hline
Datasets & Task & CCA & CFA & {\begin{tabular}{c} KCCA\\(Poly) \end{tabular}} & {\begin{tabular}{c} KCCA\\(RBF) \end{tabular}} & {\begin{tabular}{c} Bimodal\\AE \end{tabular}} & {\begin{tabular}{c} Multimodal\\DBN \end{tabular}} & Corr-AE & JRL & CMDN & \textbf{CDMLMR}\\
\hline
\multirow{3}{*}{\begin{tabular}{c} Wikipedia \\ Dataset \end{tabular}} &
Image$\rightarrow$Text & 0.186 &0.315 &0.245 & 0.275 & 0.282 & 0.189 & 0.335 & 0.310 & 0.360  &\textbf{0.388} \\
\cline{2-12}
& Text$\rightarrow$Image & 0.167 & 0.328 & 0.277 & 0.341 & 0.327 & 0.222 & 0.368 & 0.386 & 0.487 & \textbf{0.517} \\
\cline{2-12}
& Average & 0.177 & 0.322 & 0.261 & 0.308 & 0.305 & 0.206 & 0.352 & 0.348 &  0.424 & \textbf{0.453} \\
\hline
\multirow{3}{*}{\begin{tabular}{c} NUS-WDIE\\-10k Dataset \end{tabular}} &
Image$\rightarrow$Text & 0.205 &0.324 &0.254 & 0.301 & 0.250 & 0.173 & 0.331 & 0.348 & 0.432 & \textbf{0.487} \\
\cline{2-12}
& Text$\rightarrow$Image & 0.210 & 0.332 & 0.250 & 0.360 & 0.297 & 0.203 & 0.379 & 0.458 & 0.497 & \textbf{0.553} \\
\cline{2-12}
& Average & 0.208 & 0.328 & 0.252 & 0.331 & 0.274 & 0.188 & 0.355 & 0.403 & 0.465 & \textbf{0.520} \\
\hline
\end{tabular} 
\end{center}
\vspace{-3mm}
\label{table:res_wiki_50}
\end{table*}

\subsection{Details of the Network}

The network parameters need to be adjusted according to the input dimensions. Here we will present the layer parameters designed for Wikipedia dataset which will be introduced in the experiment section.
In the base network, the two-layer DBN for image input has 2048 hidden units on the first layer, and on the second layer, there are 1024 hidden units. For the text input, the two-layer DBN has 1024 hidden units on both the two layers. On the top of DBN, a three-layer feed-forward neural network with a Softmax layer is adopted for further optimization, which has the dimensional number of 1024 on each layer. In the Bimodal AE, the input layer and the reconstruction layer have the same number of dimension, and the dimensional number of the middle layer is half of the input. There is also a Softmax layer connected to the middle layer for further optimization. As for the two-pathway network in Figure \ref{fig:network}, all the three fully-connected layers have the dimensional number of 256. And the dimension of the fully-connected layers with sigmoid nonlinearity on each loss branch is also 256. For generality, the output dimensions are 256 for all the 3 datasets according to the retrieval accuracy on validation set of Wikipedia dataset. The networks are trained with a base learning rate 0.001 by stochastic gradient descent with 0.9 momentum, and the weight decay parameter is 0.004. The network is easy to train and converges in less than 5k steps in our experiment. 


\section{Experiment}

\subsection{Experiment Datasets}
We will introduce the 2 datasets briefly as follows. For fair comparison, the dataset partition and feature extraction is strictly the same with \cite{feng12014cross} and \cite{DBLP:conf/ijcai/PengHQ16}. It should be noted that in our experiment, unlabeled data is from test set, so we set the ratio of labeled/unlabeled data according to the ratio of training/test data.

{\textbf {Wikipedia dataset}} \cite{RasiwasiaMM10SemanticCCA}. Based on Wikipedia's ``feature articles", Wikipedia dataset is the most widely used dataset for cross-modal retrieval, which consists of 2,866 documents with 10 categories, and each document has an image/text pair. 
In our experiment, following \cite{feng12014cross} and \cite{DBLP:conf/ijcai/PengHQ16}, the dataset is  split into 3 parts: 2,173 documents as training set, 462 documents as testing set, and 231 documents as validation set. The image feature is 2,296-d catenation of 1,000-d PHOW descriptor, 512-d GIST descriptor, and 784-d MPEG-7 descriptor. The text feature is the representation of 3,000-d BoW vector.


{\textbf {NUS-WIDE-10k dataset}} \cite{NUSWIDE}. NUS-WIDE dataset consists of about 270,000 images and the tags of them, and they are categorized into 81 classes.  NUS-WIDE-10k is a subset of NUS-WIDE dataset with 10,000 image/text pairs from the 10 largest classes (1,000 image/pairs from each class). The dataset is also randomly split into 3 parts: 8,000 documents as training set, 1,000 documents as testing set, and 1,000 documents as validation set. The same as \cite{feng12014cross}, we take 1,134-d catenation image feature of 64-d color histogram, 144-d color correlogram, 73-d edge direction histogram, 128-d wavelet texture, 225-d block-wise color moments and 500-d SIFT-based BoVW features. The texts are represented by 1,000-d BoW vector.


\subsection{Compared Methods and Evaluation Metric}

Two retrieval tasks are conducted: retrieving text by image query (Image$\rightarrow$Text) and retrieving image by text query (Text$\rightarrow$Image), where each image in test set is used to retrieve all the text in the test set and vice versa. 8 state-of-the-art cross-modal retrieval methods are used for comparison: CCA \cite{HotelingBiometrika36RelationBetweenTwoVariates}, CFA \cite{LiMM03CFA}, KCCA \cite{DBLP:journals/neco/HardoonSS04} (with Poly and RBF kernel functions), Bimodal AE \cite{ngiam32011multimodal}, Multimodal DBN \cite{srivastava2012learning}, Corr-AE \cite{feng12014cross}, JRL \cite{ZhaiTCSVT2014JRL} and CMDN \cite{DBLP:conf/ijcai/PengHQ16}. After obtaining cross-modal shared representations by CDMLMR and the compared methods, we get the ranking list with cosine distance and adopt mean average precision (MAP) score as evaluation metric for both \emph{all} and \emph{top 50} results.

\subsection{Experimental Results}

Table \ref{table:res_wiki_All} and \ref{table:res_wiki_50} show the MAP scores on the 2 datasets for all and top 50 results. We can see on both Wikipedia and NUS-WIDE datasets, CDMLMR achieves inspiring improvement compared with the state-of-the-art methods in Image$\rightarrow$Text and Text$\rightarrow$Image tasks. In general, KCCA shows clear advantage than CCA because of its non-linearity, and JRL achieves the hight accuracy in methods without DNN. As for DNN-based methods, CMDN has the best performance in the four DNN-based compared methods (Bimodal AE, Multimodal DBN, Corr-AE and CMDN) because it models the inter-modal and intra-modal information simultaneously. As shown from the above results, our CDMLMR method can measure the cross-modal similarity more effectively.
Compared to the existing methods, CDMLMR can fully capture the useful yet intrinsic hints for cross-modal similarity metric by exploiting the similarity ranking information of cross-modal quadruplets, and make full use of unlabeled data to increase the diversity of training data. Thus we can learn  semantically discriminative shared representations and boost the cross-modal retrieval accuracy.

Table \ref{table:Baseline} shows the experiments of our CDMLMR method and the base network baselines. We compared three baselines: \textbf{Base} means to directly perform retrieval with the output of base network; \textbf{Semi} means to only use the semi-supervised contrastive loss; \textbf{Quad} means to only use the quadruplet ranking loss. For the page limitation, we only show the MAP score for \emph{all results} here.
It can be seen that the CDMLMR clearly improves the cross-modal retrieval accuracy, which shows that the similarity ranking information of cross-modal quadruplets and the rich cross-modal unlabeled instances provide useful hints for cross-modal similarity learning, and can be effectively modeled by our CDMLMR model in a unified framework. 

\begin{table}[!htb]
\scriptsize
\caption{The MAP scores for \emph{all results} of baselines.}
\vspace{-3mm}
\begin{center}
\begin{tabular}{|c|c|c|c|c|c|c|c|c|}
\hline \multicolumn{1}{|c|}{Dataset}
&\multicolumn{1}{c|}{Task}
&\multicolumn{1}{c|}{Base}  &\multicolumn{1}{c|}{Semi}  &\multicolumn{1}{c|}{Quad} &\multicolumn{1}{c|}{CDMLMR} \\  \hline

\multirow {3}*{\begin{tabular}{c}Wikipedia   \\dataset\end{tabular}}
&\multicolumn{1}{c|}{ Image$\rightarrow$Text}           &\multicolumn{1}{c|}{0.292}    &\multicolumn{1}{c|}{ 0.364}  &\multicolumn{1}{c|}{ 0.344}    &\multicolumn{1}{c|}{ \textbf{0.412}}   \\
\cline{2-6}

&\multicolumn{1}{c|}{ Text$\rightarrow$Image} &\multicolumn{1}{c|}{0.240}    &\multicolumn{1}{c|}{ 0.308}  &\multicolumn{1}{c|}{ 0.280}    &\multicolumn{1}{c|}{ \textbf{0.341}}    \\
\cline{2-6}

&\multicolumn{1}{c|}{ Average}  &\multicolumn{1}{c|}{0.266}    &\multicolumn{1}{c|}{ 0.336}  &\multicolumn{1}{c|}{ 0.312}    &\multicolumn{1}{c|}{ \textbf{0.377}}  \\ \hline

\multirow {3}*{\begin{tabular}{c}NUS-WIDE\\ -10k dataset\end{tabular}}
&\multicolumn{1}{c|}{ Image$\rightarrow$Text}         &\multicolumn{1}{c|}{0.264}    &\multicolumn{1}{c|}{ 0.351}  &\multicolumn{1}{c|}{ 0.326}    &\multicolumn{1}{c|}{ \textbf{0.405}}   \\
\cline{2-6}

&\multicolumn{1}{c|}{ Text$\rightarrow$Image}	&\multicolumn{1}{c|}{0.290}    &\multicolumn{1}{c|}{ 0.342}  &\multicolumn{1}{c|}{ 0.312}    &\multicolumn{1}{c|}{ \textbf{0.379}}     \\
\cline{2-6}

&\multicolumn{1}{c|}{ Average}  &\multicolumn{1}{c|}{0.277}    &\multicolumn{1}{c|}{ 0.347}  &\multicolumn{1}{c|}{ 0.319}    &\multicolumn{1}{c|}{ \textbf{0.392}}   \\ \hline




\end{tabular}
\end{center}
\vspace{-5mm}
\label{table:Baseline}
\end{table}

\section{Conclusion}
This paper has proposed CDMLMR modal which integrates quadruplet ranking loss and semi-supervised contrastive loss for modeling cross-modal semantic similarity in a unified multi-task learning architecture. 
Compared to the existing methods, CDMLMR can not only exploit the similarity ranking information of cross-modal quadruplets to learn the semantically discriminative shared representations, but also make full use of unlabeled data to increase the diversity of training data, and can improve the retrieval accuracy.
In the future, we still focus on cross-modal deep metric learning and aim at modeling more than two modalities simultaneously. 

\section{Acknowledgments}
This work was supported by National Natural Science Foundation of China under Grants 61371128 and 61532005.

\bibliographystyle{IEEEbib}
\bibliography{camera-ready_icme2017template}

\begin{thebibliography}{10}

\bibitem{HotelingBiometrika36RelationBetweenTwoVariates}
Harold Hotelling,
\newblock ``Relations between two sets of variates,''
\newblock {\em Biometrika}, pp. 321--377, 1936.

\bibitem{RasiwasiaMM10SemanticCCA}
Nikhil Rasiwasia, Jose Costa~Pereira, Emanuele Coviello, Gabriel Doyle, Gert~RG
  Lanckriet, Roger Levy, and Nuno Vasconcelos,
\newblock ``A new approach to cross-modal multimedia retrieval,''
\newblock in {\em ACM International Conference on Multimedia (ACM-MM)}, 2010,
  pp. 251--260.

\bibitem{DBLP:journals/ijcv/GongKIL14}
Yunchao Gong, Qifa Ke, Michael Isard, and Svetlana Lazebnik,
\newblock ``A multi-view embedding space for modeling internet images, tags,
  and their semantics,''
\newblock {\em International Journal of Computer Vision (IJCV)}, vol. 106, no.
  2, pp. 210--233, 2014.

\bibitem{LiMM03CFA}
Dongge Li, Nevenka Dimitrova, Mingkun Li, and Ishwar~K Sethi,
\newblock ``Multimedia content processing through cross-modal association,''
\newblock in {\em ACM International Conference on Multimedia (ACM-MM)}, 2003,
  pp. 604--611.

\bibitem{ZhaiTCSVT2014JRL}
Xiaohua Zhai, Yuxin Peng, and Jianguo Xiao,
\newblock ``Learning cross-media joint representation with sparse and
  semi-supervised regularization,''
\newblock {\em IEEE Transactions on Circuits and Systems for Video Technology
  (TCSVT)}, vol. 24, pp. 965--978, 2014.

\bibitem{DBLP:journals/tmm/KangXLXP15}
Cuicui Kang, Shiming Xiang, Shengcai Liao, Changsheng Xu, and Chunhong Pan,
\newblock ``Learning consistent feature representation for cross-modal
  multimedia retrieval,''
\newblock {\em IEEE Transactions on Multimedia (TMM)}, vol. 17, no. 3, pp.
  370--381, 2015.

\bibitem{DBLP:journals/tmm/HuaWLCH16}
Yan Hua, Shuhui Wang, Siyuan Liu, Anni Cai, and Qingming Huang,
\newblock ``Cross-modal correlation learning by adaptive hierarchical semantic
  aggregation,''
\newblock {\em IEEE Transactions on Multimedia (TMM)}, vol. 18, no. 6, pp.
  1201--1216, 2016.

\bibitem{ngiam32011multimodal}
Jiquan Ngiam, Aditya Khosla, Mingyu Kim, Juhan Nam, Honglak Lee, and Andrew~Y
  Ng,
\newblock ``Multimodal deep learning,''
\newblock in {\em International Conference on Machine Learning (ICML)}, 2011,
  pp. 689--696.

\bibitem{srivastava42012multimodal}
Nitish Srivastava and Ruslan Salakhutdinov,
\newblock ``Multimodal learning with deep boltzmann machines,''
\newblock in {\em Conference on Neural Information Processing Systems (NIPS)},
  2012, pp. 2222--2230.

\bibitem{DBLP:conf/ijcai/PengHQ16}
Yuxin Peng, Xin Huang, and Jinwei Qi,
\newblock ``Cross-media shared representation by hierarchical learning with
  multiple deep networks,''
\newblock in {\em International Joint Conference on Artificial Intelligence
  (IJCAI)}, 2016, pp. 3846--3853.

\bibitem{DBLP:conf/icml/AndrewABL13}
Galen Andrew, Raman Arora, Jeff~A. Bilmes, and Karen Livescu,
\newblock ``Deep canonical correlation analysis,''
\newblock in {\em International Conference on Machine Learning (ICML)}, 2013,
  pp. 1247--1255.

\bibitem{DBLP:conf/cvpr/YanM15}
Fei Yan and Krystian Mikolajczyk,
\newblock ``Deep correlation for matching images and text,''
\newblock in {\em Conference on Computer Vision and Pattern Recognition
  (CVPR)}, 2015, pp. 3441--3450.

\bibitem{feng12014cross}
Fangxiang Feng, Xiaojie Wang, and Ruifan Li,
\newblock ``Cross-modal retrieval with correspondence autoencoder,''
\newblock in {\em ACM International Conference on Multimedia (ACM-MM)}, 2014,
  pp. 7--16.

\bibitem{DBLP:conf/cvpr/WangSLRWPCW14}
Jiang Wang, Yang Song, Thomas Leung, Chuck Rosenberg, Jingbin Wang, James
  Philbin, Bo~Chen, and Ying Wu,
\newblock ``Learning fine-grained image similarity with deep ranking,''
\newblock in {\em Conference on Computer Vision and Pattern Recognition
  (CVPR)}, 2014, pp. 1386--1393.

\bibitem{DBLP:conf/cvpr/LaiPLY15}
Hanjiang Lai, Yan Pan, Ye~Liu, and Shuicheng Yan,
\newblock ``Simultaneous feature learning and hash coding with deep neural
  networks,''
\newblock in {\em Conference on Computer Vision and Pattern Recognition
  (CVPR)}, 2015, pp. 3270--3278.

\bibitem{DBLP:conf/nips/RenHGS15}
Shaoqing Ren, Kaiming He, Ross~B. Girshick, and Jian Sun,
\newblock ``Faster {R-CNN:} towards real-time object detection with region
  proposal networks,''
\newblock in {\em Conference on Neural Information Processing Systems (NIPS)},
  2015, pp. 91--99.

\bibitem{NUSWIDE}
Tat{-}Seng Chua, Jinhui Tang, Richang Hong, Haojie Li, Zhiping Luo, and Yantao
  Zheng,
\newblock ``Nus-wide: a real-world web image database from national university
  of singapore,''
\newblock in {\em ACM International Conference on Image and Video Retrieval
  (CIVR)}, 2009.

\bibitem{DBLP:journals/neco/HardoonSS04}
David~R. Hardoon, S{\'{a}}ndor Szedm{\'{a}}k, and John Shawe{-}Taylor,
\newblock ``Canonical correlation analysis: An overview with application to
  learning methods,''
\newblock {\em Neural Computation}, vol. 16, no. 12, pp. 2639--2664, 2004.

\bibitem{srivastava2012learning}
Nitish Srivastava and Ruslan Salakhutdinov,
\newblock ``Learning representations for multimodal data with deep belief
  nets,''
\newblock in {\em International Conference on Machine Learning (ICML)
  Workshop}, 2012.

\end{thebibliography}

\end{document}